\documentclass[10pt,twocolumn,letterpaper]{article}

\usepackage[pagenumbers]{cvpr} 

\usepackage{graphicx}
\usepackage{amsmath}
\usepackage{amssymb}
\usepackage{booktabs}

\usepackage[dvipsnames]{xcolor}
\usepackage{makecell}
\usepackage[pagebackref,breaklinks,colorlinks]{hyperref}

\usepackage[accsupp]{axessibility} 

\definecolor{mypink}{RGB}{255,125,214}
\definecolor{mygreen}{RGB}{55, 192, 120}
\definecolor{myyellow}{RGB}{255,192,0}
\definecolor{myorange}{RGB}{255,115, 90}
\definecolor{myblue}{RGB}{80, 130, 255}
\definecolor{openred}{RGB}{255, 90, 96}
\definecolor{openblue}{RGB}{104, 158, 184}
\definecolor{blockblue}{RGB}{157, 195, 230}
\definecolor{blockorange}{RGB}{244, 177, 131}

\usepackage[capitalize]{cleveref}
\crefname{section}{Sec.}{Secs.}
\Crefname{section}{Section}{Sections}
\Crefname{table}{Table}{Tables}
\crefname{table}{Tab.}{Tabs.}

\newcommand{\egoloss}{$\mathcal{L}_{ego}^{\textcolor{green}{\bullet}}$}
\newcommand{\segloss}{$\mathcal{L}_{seg}^{\textcolor{cyan}{\bullet}}$}
\newcommand{\flowloss}{$\mathcal{L}_{flow}^{\textcolor{orange}{\bullet}}$}

\def\cvprPaperID{9751} 
\def\confName{CVPR}
\def\confYear{2023}

\begin{document}

\title{Hidden Gems: 4D Radar Scene Flow Learning Using \\Cross-Modal Supervision }

\author{Fangqiang Ding$^{1}$
\and
Andras Palffy$^{2}$
\and
Dariu M. Gavrila$^{2}$
\and
Chris Xiaoxuan Lu$^{1,}$\thanks{Corresponding author: Chris Xiaoxuan Lu ({xiaoxuan.lu@ed.ac.uk})}
\and 
$^{1}$University of Edinburgh
\\
{\tt\small \{fding, xiaoxuan.lu\}@ed.ac.uk}
\and
$^{2}$Delft University of Technology
\\
{\tt\small \{a.palffy, d.m.gavrila\}@tudelft.nl}
}
\maketitle

\begin{abstract}
This work proposes a novel approach to 4D radar-based scene flow estimation via cross-modal learning. Our approach is motivated by the co-located sensing redundancy in modern autonomous vehicles. Such redundancy implicitly provides various forms of supervision cues to the radar scene flow estimation. Specifically, we introduce a multi-task model architecture for the identified cross-modal learning problem and propose loss functions to opportunistically engage scene flow estimation using multiple cross-modal constraints for effective model training. Extensive experiments show the state-of-the-art performance of our method and demonstrate the effectiveness of cross-modal supervised learning to infer more accurate 4D radar scene flow. We also show its usefulness to two subtasks - motion segmentation and ego-motion estimation. Our source code will be available on {\href{https://github.com/Toytiny/CMFlow}{https://github.com/Toytiny/CMFlow}}.
\end{abstract}

\vspace{-1em}
\section{Introduction}

Scene flow estimation is to obtain a 3D motion vector field of the static and dynamic environment relative to an ego-agent. In the context of self-driving, scene flow is a key enabler to navigational safety in dynamic environments by providing holistic motion cues to multiple subtasks, such as ego-motion estimation, motion segmentation, point cloud accumulation,  multi-object tracking, \etc.

Driven by the recent successes of deep neural networks in point cloud processing~\cite{qi2017pointnet,qi2017pointnet++,wu2019pointconv,su2018splatnet, zhou2018voxelnet, lang2019pointpillars}, predominant approaches to scene flow estimation from point clouds adopt either fully-~\cite{liu2019flownet3d,wu2020pointpwc,puy2020flot,gu2019hplflownet,wei2021pv} or weakly-~\cite{gojcic2021weakly,dong2022exploiting} supervised learning, or only rely on self-supervised signals~\cite{li2022rigidflow, baur2021slim, mittal2020just,ding2022raflow,kittenplon2021flowstep3d}. For supervised ones, the acquisition of scene flow annotations is costly and requires tedious and intensive human labour. In contrast, self-supervised learning methods require no annotations and can exploit the inherent spatio-temporal relationship and constraints in the input data to bootstrap scene flow learning. Nevertheless, due to the implicit supervision signals, self-supervised learning performance is often secondary to the supervised ones~\cite{wei2021pv,li2021hcrf, dong2022exploiting}, and they fail to provide sufficiently reliable results for safety-critical autonomous driving scenarios. 

\begin{figure}[tbp!]
    \centering
    \includegraphics[width=0.475\textwidth]{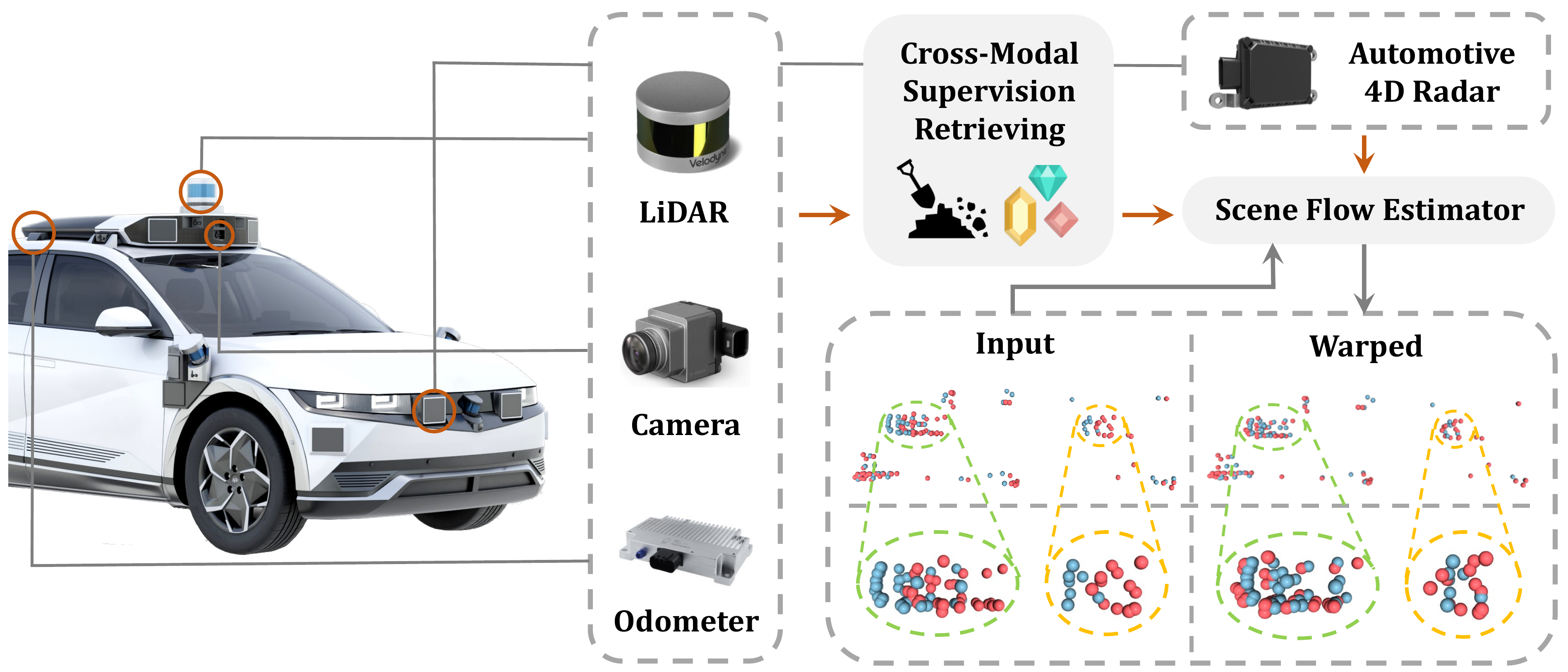}
    \caption{Cross-modal supervision cues are retrieved from co-located odometer, LiDAR and camera sensors to benefit 4D radar scene flow learning. The source point cloud (\textcolor{openred}{red}) is warped with our estimated scene flow and gets closer to the target one (\textcolor{openblue}{blue}). }
    \label{fig:open}
    \vspace{-1em}
\end{figure}

These challenges become more prominent when it comes to 4D radar scene flow learning. 4D automotive radars receive increasing attention recently due to their robustness against adverse weather and poor lighting conditions (\vs camera), availability of object velocity measurements (\vs camera and LiDAR) and relatively low cost (\vs LiDAR)~\cite{palffy2022multi, sun20214d, brisken2018recent, meyer2019automotive}.
However, 4D radar point clouds are significantly sparser than LiDARs' and suffer from non-negligible multi-path noise. Such low-fidelity data significantly complicates the point-level scene flow annotation for supervised learning and makes it difficult to rely exclusively on self-supervised training-based methods \cite{ding2022raflow} for performance and safety reasons. 

To find a more effective framework for 4D radar scene flow learning, this work aims to exploit cross-modal supervision signals in autonomous vehicles. Our motivation is based on the fact that autonomous vehicles today are equipped with multiple heterogeneous sensors, \eg, LiDARs, cameras and GPS/INS, which can provide complementary sensing and redundant perception results for each other, jointly safeguarding the vehicle operation in complex urban traffic. This \emph{co-located perception redundancy} can be leveraged to provision multiple supervision cues that bootstrap radar scene flow learning. For example, the static points identified by radar scene flow can be used to estimate the vehicle odometry. The consistency between this estimated odometry and the observed odometry from the co-located GPS/INS on the vehicle forms a natural constraint and can be used to supervise scene flow estimation. Similar consistency constraints can be also found for the optical flow estimated from co-located cameras and the projected radar scene flow onto the image plane. 

While the aforementioned examples are intuitive, retrieving accurate supervision signals from co-located sensors is non-trivial. With the same example of optical and scene flow consistency, minimizing flow errors on the image plane suffers from the depth-unaware perspective projection, potentially incurring weaker constraints to the scene flow of far points. This motivates the following research question: \emph{How to retrieve the cross-modal supervision signals from co-located sensors on a vehicle and apply them collectively to bootstrap radar scene flow learning}.


Towards answering it, in this work we consider \emph{opportunistically} exploiting useful supervision signals from three commonly co-existent sensors with the 4D radar on a vehicle: odometer (GPS/INS), LiDAR, and RGB camera (See~\cref{fig:open}). This cross-modal supervision is expected to help us realize radar scene flow learning without human annotation. In our setting, the multi-modal data are only available in the training phase, and only 4D radar is used during the inference stage. Our contributions can be summarized as follows:
\vspace{-0.5em}

\begin{itemize}
    \setlength{\itemsep}{0pt}
    \setlength{\parsep}{0pt}
    \setlength{\parskip}{0pt}
    \item Our work is the first 4D radar scene flow learning using cross-modal supervision from co-located heterogeneous sensors on an autonomous vehicle.  
    \item We introduce a multi-task model architecture for the identified cross-modal learning problem and propose loss functions to effectively engage scene flow estimation using multiple cross-modal constraints for model training. 
    \item We demonstrate the state-of-the-art performance of the proposed CMFlow method on a public dataset and show its effectiveness in downstream tasks as well.
\end{itemize}

\section{Related Work}
\label{sec:related}

\noindent\textbf{Scene flow.} Scene flow was first defined in~\cite{vedula2005three} as a 3D uplift of optical flow that describes the 3D displacement of points in the scene.  
Traditional approaches resolve pixel-wise scene flow from either RGB
or {RGB-D}\linebreak images
based on prior knowledge assumptions~\cite{huguet2007variational,wedel2008efficient,vcech2011scene,vogel2013piecewise, hornacek2014sphereflow, quiroga2013local, quiroga2014dense, hadfield2011kinecting} or by training deep networks in a supervised~\cite{mayer2016large, ilg2018occlusions, lv2018learning, jiang2019sense, qiao2018sfnet} or unsupervised~\cite{liu2019unsupervised, yin2018geonet, zou2018df, hur2020self} way. In contrast, some other methods directly infer point-wise scene flow from 3D sparse point clouds. Among them, some methods rely on online optimization to solve scene flow~\cite{dewan2016rigid,pontes2020scene,li2021neural}. Recently, inspired by the success of point cloud feature learning~\cite{qi2017pointnet, qi2017pointnet++,wu2019pointconv,su2018splatnet}, deep learning-based methods~\cite{behl2019pointflownet,liu2019flownet3d,gu2019hplflownet,wu2020pointpwc} have been dominant for point cloud-based scene flow estimation. 

\noindent\textbf{Deep scene flow on point clouds.} Current point cloud-based scene flow estimation methods~\cite{behl2019pointflownet,liu2019flownet3d,gu2019hplflownet,wu2020pointpwc,li2022rigidflow, puy2020flot,kittenplon2021flowstep3d,gojcic2021weakly,mittal2020just,baur2021slim,wei2021pv,dong2022exploiting,wang2020flownet3d++,wang2021festa} established state-of-the-art performance by leveraging large amount of data for training. Many of them~\cite{wang2020flownet3d++,wang2021festa,puy2020flot,wei2021pv,behl2019pointflownet} learn scene flow estimation in a fully-supervised manner with ground truth flow. These methods, albeit showing promising results, demand scene flow annotations, of which the acquisition is labour-intensive and costly. Another option is to leverage a simulated dataset~\cite{mayer2016large} for training, yet this may result in poor generalization when applied to the real data. To avoid both human labour and the pitfalls of synthetic data, some methods design self-supervised learning frameworks~\cite{kittenplon2021flowstep3d,li2022rigidflow,baur2021slim, mittal2020just,wu2020pointpwc} that exploit supervision signals from the input data. Compared with their supervised counterparts, these methods require no annotated labels and thus can train their models on unannotated datasets. However, the performance of these self-supervised learning methods is limited~\cite{wu2020pointpwc, kittenplon2021flowstep3d} by the fact that no real labels are used to supervise their models. 
Some recent methods~\cite{gojcic2021weakly,dong2022exploiting} try to seek a trade-off between annotation efforts and performance by combining the ego-motion and manually annotated background segmentation labels.
Although \emph{pseudo} ground truth ego-motion can be easily accessed from onboard odometry sensors (GPS/INS), annotating background mask labels is still expensive and need human experts to identify foreground entities from complex scenarios. 

\noindent\textbf{Radar scene flow.}
Previous works mostly estimate scene flow on dense point clouds captured by LiDAR or rendered from stereo images. Thus, they cannot be directly extended to the sparse and noisy radar point clouds. To fill the gap, a recent work~\cite{ding2022raflow} proposes a self-supervised pipeline for radar scene flow estimation. However, just like in other self-supervised methods, the lack of real supervision signals limits its scene flow estimation performance and thus hinders its application to more downstream tasks. 

Unlike the aforementioned works, we propose to retrieve supervision signals from co-located sensors in an automatic manner without resorting to any human intervention during training. Note that we use cross-modal data to provide all supervision at once only in the training stage and do not require other modalities during inference.

\section{Method}\label{sec:method}
\subsection{Problem Definition}
Scene flow estimation aims to solve a motion field that describes the non-rigid transformations induced both by the motion of the ego-vehicle and the dynamic objects in the scene. For point cloud-based scene flow, the inputs are two consecutive point clouds, the source one $\mathbf{P}^{s}=\{\mathbf{p}^s_i = \{\mathbf{c}^s_i, \mathbf{x}^s_i\}\}_{i=1}^N$ and the target one $\mathbf{P}^{t}=\{\mathbf{p}^t_i = \{\mathbf{c}^t_i, \mathbf{x}^t_i\}\}_{i=1}^M$, where $\mathbf{c}^s_i$, $\mathbf{c}^t_i\in\mathbb{R}^3$ are the 3D coordinates of each point, and $\mathbf{x}^s_i$, $\mathbf{x}^t_i\in\mathbb{R}^C$ are their associated raw point features. The outputs are point-wise 3D motion vectors $\mathbf{F}=\{\mathbf{f}_i\in\mathbb{R}^3\}_{i=1}^N$ that align each point in $\mathbf{P}^s$ to its corresponding position $\mathbf{c}'_i = \mathbf{c}^s_i+ \mathbf{f}_i$ in the target frame. Note that $\mathbf{P}^s$ and $\mathbf{P}^t$ do not necessarily have the same number of points and there is no strict point-to-point correspondence between them under real conditions. Therefore, the corresponding location $\mathbf{c}'_i$ is not required to coincide with any points in the target point cloud $\mathbf{P}^t$. In our case of 4D radar, the raw point features  $\mathbf{x}^s_i$, $\mathbf{x}^t_i\in\mathbb{R}^2$ include the relative radial velocity (RRV) and the radar cross-section (RCS) measurements~\cite{barton2004radar}. RRV measurements, resolved by analyzing Doppler shift in radar frequency data, contain partial motion information of points. RCS can be seen as the proxy reflectivity of each point, which is mainly affected by the reflection property of the target and the incident angle of the beams. 

\begin{figure*}[tbp!]
    \centering
    \includegraphics[width=0.975\textwidth]{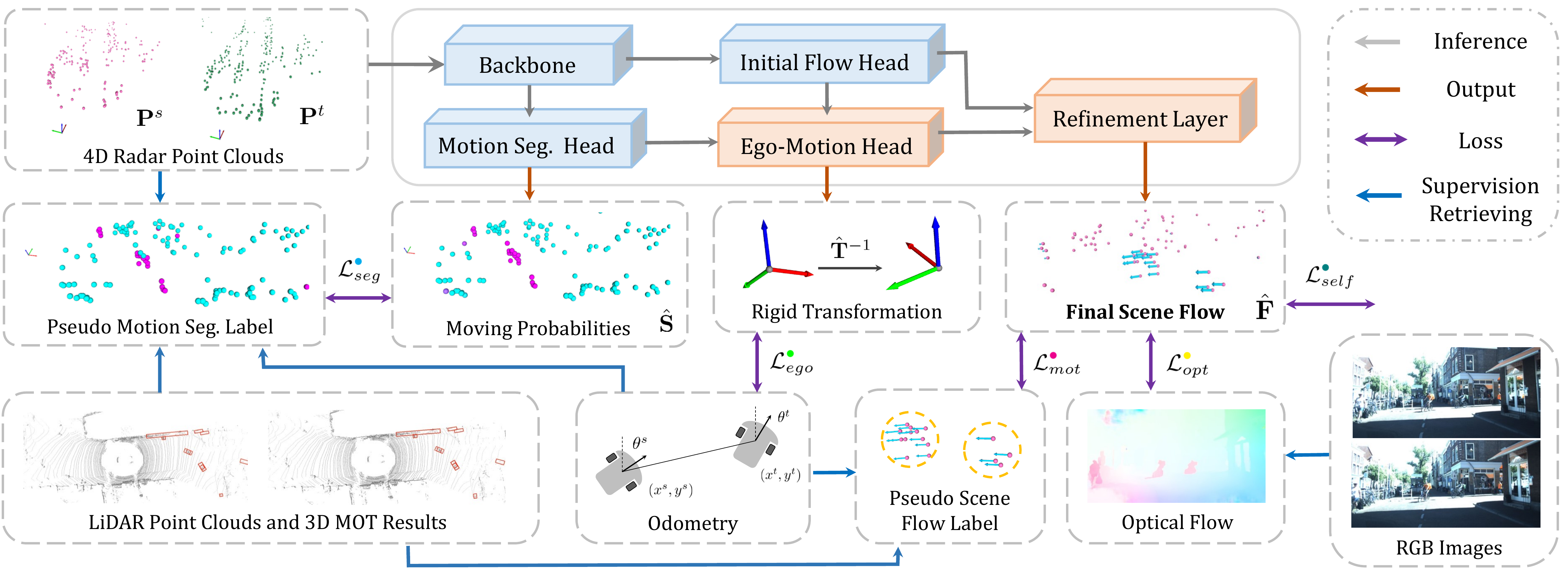}
    \caption{Cross-modal supervised learning pipeline for 4D radar scene flow estimation. The model architecture (c.f.~\cref{sec:model}) is composed of two stages (\textcolor{blockblue}{blue}/\textcolor{blockorange}{orange} block colours for stage 1/2) and outputs the final scene flow together with the motion segmentation and a rigid ego-motion transformation. Cross-modal supervision signals are utilized to constrain outputs with various loss functions (c.f.~\cref{sec:loss}).} 
    \label{fig:pipeline}
    \vspace{-1em}
\end{figure*}

\subsection{Overview}

The overall pipeline of our proposed method is depicted in~\cref{fig:pipeline}. To bootstrap cross-modal supervised learning, we apply a multi-task model that predicts radar scene flow in two stages.
The first stage starts by extracting base features with two input point clouds, which are then forwarded to two independent heads to infer per-point moving probabilities and initial point-wise scene flow vectors. On top of them, the second stage first infers a rigid transformation respective to radar ego-motion and outputs a binary motion segmentation mask. Then, the final scene flow is obtained by refining the flow vectors of identified static points with the derived rigid transformation. 
In summary, our multi-task model's outputs include a rigid transformation (\ie the ego-motion), a motion segmentation mask (\ie which targets are static or dynamic), and the final refined scene flow. 

To supervise these predictions, we extract corresponding supervision signals from co-located modalities and train the entire model end-to-end by minimizing a loss $\mathcal{L}$ composed of three terms:
\begin{equation}\label{eq:loss}\small
    \mathcal{L} =  \mathcal{L}_{ego}^{\textcolor{green}{\bullet}} +  \mathcal{L}_{seg}^{\textcolor{cyan}{\bullet}} + \mathcal{L}_{flow}^{\textcolor{orange}{\bullet}}~.
\end{equation}
Here, \egoloss~is the ego-motion error that supervises the rigid transformation estimation using the odometry information. Our motion segmentation error \segloss~constrains the predicted moving probabilities with a point-wise pseudo motion segmentation label, which is obtained by fusing information from the odometer and LiDAR. We further supervise the final scene flow with signals given by LiDAR and RGB camera in \flowloss.
In the following section (\cref{sec:model}), we first briefly introduce each module of our model.
We then explain how we extract signals from co-located modalities to supervise our outputs in the training phase (\cref{sec:loss}). More details on our method can be found in the supplementary materials.

\subsection{Model Architecture}\label{sec:model}
Similar to~\cite{tishchenko2020self,baur2021slim,ding2022raflow,dong2022exploiting, gojcic2021weakly}, our model is designed in a two-stage fashion, with a rough initial flow derived in the first stage and refined in the second stage to obtain the final estimate of scene flow, as shown in~\cref{fig:pipeline}. A description of all components of our model can be found below.

\noindent\textbf{Backbone.}
Following~\cite{liu2019flownet3d,wu2020pointpwc,dong2022exploiting, ding2022raflow}, our backbone network directly operates on unordered point clouds  $\mathbf{P}^{s}$ and $\mathbf{P}^{t}$ to encode point-wise latent features. In particular, we first apply the \emph{set conv} layers~\cite{liu2019flownet3d} to robustly extract multi-scale local features for individual point clouds and propagate features from $\mathbf{P}^t$ to $\mathbf{P}^s$ using the~\emph{cost volume} layer~\cite{wu2020pointpwc} for feature correlation. We then concatenate multi-stage features of $\mathbf{P}^s$ (including the correlated features) and forward them into another multi-scale $\emph{set conv}$ layers to generate the base backbone features $\mathbf{E}\in\mathbb{R}^{N\times C_e}$. Specifically, the {max-pooling} operation is used along the channel axis to restore the global scene features after each \emph{set conv} layer, which are then concatenated to per-point local features.

\noindent\textbf{Initial flow and motion segmentation head.} Given the base backbone features $\mathbf{E}$ that represents the inter-frame motion and intra-frame local-global information for each point $\mathbf{p}^s_{i}\in\mathbf{P}^s$, we apply two task-specific heads for decoding. The first head is to produce an initial scene flow $\hat{\mathbf{F}}^{init}=\{\hat{\mathbf{f}}_i^{init}\in\mathbb{R}^3\}_{i=1}^N$, while another is for generating a probability map $\hat{\mathbf{S}}=\{\hat{s}_i\in[0,1]\}_{i=1}^N$ that denotes the moving probabilities of points in $\mathbf{P}^s$ (\emph{w.r.t.} the world frame). We implement both heads with a multi-layer perceptron and map the output to probabilities with $Sigmoid (\cdot)$ in the motion segmentation head. 

\noindent\textbf{Ego-motion head.} With the natural correspondences $\{\mathbf{c}^s_i,\mathbf{c}^s_i+\hat{\mathbf{f}}_i^{init}\}_{i=1}^N$ formed by the initial scene flow $\hat{\mathbf{F}}^{init}$ between two frames and the probability map $\hat{\mathbf{S}}$\footnote{Specifically, we use the pseudo label $\bar{\mathbf{S}}$ (c.f.~\cref{sec:loss}) instead of $\hat{\mathbf{S}}$ in the ego-motion head during training for stable scene flow learning.}, we restore a rigid transformation $\hat{\mathbf{T}}\in\mathbb{R}^{4\times 4}$ that describes the radar ego-motion using the differentiable weighted Kabsch algorithm~\cite{kabsch1976solution}. To mitigate the impact of the flow vectors from moving points, we compute $1-\hat{\mathbf{S}}$ as our weights and normalize the sum of them to 1. Besides, we generate a binary motion segmentation mask by thresholding $\hat{\mathbf{S}}$ with a fixed value $\eta_b$ to indicate the moving status of each point. We use this binary mask as our motion segmentation output and to identify stationary points for flow refinement below.

\noindent\textbf{Refinement layer.} As the flow vectors of static points are only caused by the radar's ego-motion, we can regularize their initial predictions with the more reliable rigid transformation $\hat{\mathbf{T}}$. The refinement operation is simply replacing the initial flow vector $\hat{\mathbf{f}}_i^{init}$ of identified stationary points with the flow vector induced by radar's ego-motion, which is derived as $[\hat{\mathbf{f}}_i~1]^{\top}=(\hat{\mathbf{T}}-\mathbf{I}_4)[\mathbf{c}_i^s~1]^{\top}$. The final scene flow is attained as $\hat{\mathbf{F}}=\{\hat{\mathbf{f}}_i\in\mathbb{R}^3\}_{i=1}^N$.

\noindent\textbf{Temporal update module.}
Apart from the aforementioned basic modules, we also propose an optional module that can be embedded into the backbone to propagate previous latent information to the current frame. More specifically, we apply a GRU network~\cite{cho2014properties} that treats the global features in the backbone as the hidden state and update it temporally across frames. During training, we first split long training sequences into mini-clips with
a fixed length $T$ and train with mini-batches of them. The
hidden state is initialized as a zero vector for the first frame of each clip.
When evaluating on test sequences, we emulate the training
conditions by re-initializing the hidden state after $T$ frames.
\vspace{0.25em}

In general, our model can deliver solutions to three different tasks, \ie scene flow estimation, ego-motion estimation and motion segmentation, with 4D radar point clouds. Specifically, these outputs are compactly correlated to each other. For example, accurate motion segmentation results will benefit ego-motion estimation, which further largely determines the quality of the final scene flow.  

\subsection{Cross-Modal Supervision Retrieving}\label{sec:loss}
A key step in our proposed architecture is to retrieve the cross-modal supervision signals from three co-located sensors on autonomous vehicles, \ie odometer, LiDAR and camera, to support model training without human annotation. This essentially leads to a multi-task learning problem. Of course, the supervision signals from individual modality-specific tasks (\eg, optical flow) are inevitably noisier than human annotation. However, we argue that if these noisy supervision signals are well combined, then the overall noise in supervision can be suppressed and give rise to effective training anyway.
In the following, we detail how we extract cross-modal supervision signals and subtly combine them to formulate a multi-task learning problem.

\noindent\textbf{Ego-motion loss.}
To supervise the rigid transformation $\hat{\mathbf{T}}$ derived in our ego-motion head, it is intuitive to leverage the odometry information from the odometer (GPS/INS). As a key sensor for mobile autonomy, the odometer can output high-frequency ego-vehicle poses, which can be used to compute the \emph{pseudo} ground truth radar ego-motion transformation $\mathbf{O}\in\mathbb{R}^{4\times 4}$ between two frames. The ground truth rigid transformation $\mathbf{T} = \mathbf{O}^{-1}$ can then be derived to summarize the rigid flow component $\mathbf{F}^{r}=\{\mathbf{f}_{i}^{r}\}_{i=1}^{N}$ induced by the ground truth radar ego-motion, where $[\mathbf{f}_{i}^{r}~1]^{\top} = (\mathbf{T}-\mathbf{I}_4)[\mathbf{c}_i^s~1]^{\top}$. 
Our ego-motion loss is formulated as:
\begin{equation}\label{eq:ego}\small
\mathcal{L}_{ego}^{\textcolor{green}{\bullet}} = \frac{1}{N} \sum_{i=1}^{N}\Big\Vert(\mathbf{\hat{T}}-\mathbf{{T}})[\mathbf{c}_i^s~1]^{\top}\Big\Vert_2~,
\end{equation}
where we supervise the estimated $\hat{\mathbf{T}}$ by encouraging its associated rigid flow components to be close to the ground truth ones. By supervising $\hat{\mathbf{T}}$, we can implicitly constrain the initial scene flow $\hat{\mathbf{f}}_i^{init}$ for static points. More importantly, the static flow vectors in the final flow $\hat{\mathbf{F}}$ can also be supervised as the refinement is based on $\hat{\mathbf{T}}$.

\noindent\textbf{Motion segmentation loss.} Unlike ego-motion estimation, supervising motion segmentation with cross-modal data is not straightforward as no sensors provide such information. 
To utilize the odometry information, we generate a pseudo motion segmentation label with the rigid flow component $\mathbf{F}^r$ given by the odometer and the radar RRV measurements $\{{v}_i\}_{i=1}^{N}$. More specifically, we first approximate the RRV component ascribed to the radar ego-motion by $v^r_i \approx {\mathbf{u}_i^\top \mathbf{f}_i^r}/{\Delta t}$. Here, $\mathbf{u}_i$ is the unit vector with its direction pointing from the sensor to the point $\mathbf{c}^s_i$ and $\Delta t$ is time duration between two frames. Then, we compensate the radar ego-motion and get the object absolute radial velocity $\Delta v_i = \mathrm{abs}(v_i-v_i^r)$. With per-point $\Delta v_i$, the pseudo motion segmentation label $\mathbf{S}^{v}=\{s_i^{v}\in\{0,1\}\}_{i=1}^{N}$ can be derived by thresholding, where 1 denotes moving points. More details on our thresholding strategy can be found in the supplementary materials. Note that one shortcoming is that tangentially moving targets are not distinguished. 

Besides the odometry information, we also leverage the LiDAR data to generate a pseudo foreground (movable objects) segmentation label $\mathbf{S}^{fg}$. To this end, we first feed LiDAR point clouds into an off-the-shelf 3D multi-object tracking (MOT) pretrained model~\cite{weng20203d}. Then we divide radar points from $\mathbf{P}^s$ into the foreground and background using the bounding boxes (\eg pedestrian, car, cyclist) produced by 3D MOT to create $\mathbf{S}^{fg}$. Besides, we can also assign pseudo scene flow vector label to foreground points by: a) retrieving the ID for each bounding box from the first frame, b) computing the inter-frame transformation for them, c) deriving the translation vector for each inbox point based on the assumption that all points belonging to the same object share a \emph{universal} rigid transformation. The resulting pseudo scene flow label is denoted as $\mathbf{F}^{fg}=\{\mathbf{f}_i^{fg}\}_{i=1}^{N}$, where we leave the label empty for all identified background points.

Given the two segmentation labels $\mathbf{S}^{v}$ and $\mathbf{S}^{fg}$, directly fusing them is impeded by their domain discrepancy, \ie, not all foreground points are moving. Therefore, we propose to distill the moving points from $\mathbf{S}^{fg}$ by discarding foreground points that either keep still or move very slowly. We implement this by removing the rigid flow component $\mathbf{F}^r$ from $\mathbf{F}^{fg}$ and get the non-rigid flow component of all foreground points. Then we obtain a new pseudo motion segmentation label $\mathbf{S}^{l} = \{{s}_i^{l}\in\{0,1\}\}_{i=1}^{N}$ by classifying points with apparent non-rigid flow as dynamic. A more reliable pseudo motion segmentation ${\mathbf{S}}=\{{s}_i\}_{i=1}^{N}$ can be consequently obtained by fusing $\mathbf{S}^{l}$ and $\mathbf{S}^{v}$. For points classified as moving in $\mathbf{S}^{l}$, we have high confidence about their status and thus label these points as moving in ${\mathbf{S}}$. For the rest of the points, we label their ${s}_i$
according to $\mathbf{S}^{v}$. Finally, as seen in~\cref{fig:pipeline}, our motion segmentation loss can be formulated by encouraging the estimated moving probabilities $\hat{\mathbf{S}}$ to be close to the pseudo motion segmentation label ${\mathbf{S}}$:  
\begin{equation}\label{eq:seg}\small
\mathcal{L}_{seg}^{\textcolor{cyan}{\bullet}} = \frac{1}{2}( \frac{\sum_{i=1}^{N}(1-{s}_i)\mathrm{log}(1-\hat{s}_i)}{\sum_{i=1}^{N}(1-{s}_i)} +\frac{\sum_{i=1}^{N}{s}_i\mathrm{log}(\hat{s}_i)}{{\sum_{i=1}^{N}{s}_i}})~.
\end{equation}
Here, we use the average loss of static and moving losses to address the class imbalance issue.

\noindent\textbf{Scene flow loss.} In the final scene flow output $\hat{\mathbf{F}}$, the flow vectors of static points have been implicitly supervised by the ego-motion loss (c.f.~\cref{eq:ego}). In order to further constrain the flow vectors of moving points, we formulate two new loss functions. The first one is  $\mathcal{L}_{mot}^{\textcolor{magenta}{\bullet}}$, which is based on the pseudo scene flow label $\mathbf{F}^{fg}$ derived from 3D MOT results. In this loss, we only constrain the flow vectors of those moving points identified in $\mathbf{S}^l$ through:
\begin{equation}\label{loss:mot}\small
    \mathcal{L}_{mot}^{\textcolor{magenta}{\bullet}} = \frac{1}{\sum_{i=1}^{N}{s_i^l}}\sum_{i=1}^{N}\Big\Vert{s}^{l}_i(\mathbf{\hat{f}}_i-\mathbf{f}^{fg}_i)\Big\Vert_2~.
\end{equation}
In addition to utilizing the odometer and LiDAR for cross-modal supervision, we also propose to extract supervision signals from the RGB camera. Specifically, we formulate a loss $\mathcal{L}_{opt}^{\textcolor{yellow}{\bullet}}$ using pseudo optical flow labels ${\mathbf{W}}=\{{\mathbf{w}}_i\in\mathbb{R}^2\}_{i=1}^{N}$. To get this pseudo label, we first feed synchronized RGB images into a pretrained optical flow estimation model~\cite{teed2020raft} to produce an optical flow image. Then, we project the coordinate $\mathbf{c}_i^{s}$ of each point onto the image plane and draw the optical flow vector at the corresponding pixel $\mathbf{m}_i$ of each point. Given $\mathbf{W}$, it is intuitive to construct the supervision for our scene flow prediction by projecting it on the image plane. However, minimizing the flow divergence in pixel scale has less impact for far radar points due to the depth-unawareness during perspective projection. Instead, we directly take the point-to-ray distance as the training objective, which is more insensitive to points at different ranges. The loss function can be written as:
\begin{equation}\label{loss:opt}\small
    \mathcal{L}_{opt}^{\textcolor{yellow}{\bullet}} = \frac{1}{\sum_{i=1}^{N}{s}_i}\sum_{i=1}^{N}{s}_i\mathcal{D}(\mathbf{c}^s_i+\mathbf{\hat{f}}_i, \mathbf{m}_i+\mathbf{w}_i,\mathbf{\theta})
\end{equation}
where $\mathcal{D}$ denotes the operation that computes the distance between the warped point $\mathbf{c}^s_i+\mathbf{\hat{f}}_i$ and the ray traced from the warped pixel $\mathbf{m}_i+\mathbf{w}_i$. $\mathbf{\theta}$ denotes sensor calibration parameters. Note that we only consider the scene flow of moving points identified in ${\mathbf{S}}$ here. Apart from the above two loss functions used to constrain the final scene flow, we also employ the self-supervised loss $\mathcal{L}_{self}^{\textcolor{teal}{\bullet}}$ in~\cite{ding2022raflow} to complement our cross-modal supervision. See the supplementary for more details. The overall scene flow loss is formulated as:
\begin{equation}\small
    \mathcal{L}_{flow}^{\textcolor{orange}{\bullet}} = \mathcal{L}_{mot}^{\textcolor{magenta}{\bullet}} + \lambda_{opt}\mathcal{L}_{opt}^{\textcolor{yellow}{\bullet}} + \mathcal{L}_{self}^{\textcolor{teal}{\bullet}}~,
\end{equation}
where we set the weight $\lambda_{opt}=0.1$ in all our experiments.

\section{Experiments}
\label{sec:exp}

\begin{table}[tbp!]
    \renewcommand\arraystretch{1.0}
    \setlength\tabcolsep{2.2pt}
    \centering
    \resizebox{\columnwidth}{!}{%
    \begin{tabular}{@{}lccccccccccc@{}}
    \toprule
        Method & Sup. & EPE [m]$\downarrow$  &  AccS$\uparrow$ &  AccR$\uparrow$  &  RNE [m]$\downarrow$  &  MRNE [m]$\downarrow$  &  SRNE [m]$\downarrow$  \\
    \midrule 
      ICP~\cite{besl1992icp} & None & 0.344 & 0.019 & 0.106 & 0.138 & 0.148 & 0.137\\
      Graph Prior*~\cite{pontes2020scene} & None & 0.445 & 0.070 & 0.104 & 0.179 & 0.186 & 0.176\\
      JGWTF*~\cite{mittal2020just} & Self & 0.375 & 0.022 &	0.103 & 0.150 & 0.139 & 0.151 \\
      PointPWC~\cite{wu2020pointpwc} & Self & 0.422 &	0.026	& 0.113	& 0.169	& 0.154 &	0.170\\
      FlowStep3D~\cite{kittenplon2021flowstep3d}  & Self & 0.292 & 0.034	& 0.161 & 0.117 & 0.130 & 0.115\\
      SLIM*~\cite{baur2021slim} & Self & 0.323 &	0.050 &	0.170	& 0.130	& 0.151	& 0.126\\
      RaFlow~\cite{ding2022raflow} & Self & 0.226 & 0.190 &	0.390 & 0.090 & 0.114 & 0.087\\
      CMFlow & Cross & 0.141 & \textbf{0.233} & 0.499 & 0.057 & 0.073 & 0.054   \\
      CMFlow (T) & Cross & \textbf{0.130}  & 0.228	& \textbf{0.539} & \textbf{0.052} & \textbf{0.072} & \textbf{0.049} \\
    \bottomrule
    \end{tabular}
    }
    \caption{
     Scene flow evaluation results on the \emph{Test} set. 
    The mean metric values across all test frames are reported. * indicates that we reproduce these methods referring to original papers since their source codes are not public. The best results on each metric are shown in \textbf{bold}. $\uparrow$ means bigger values are better, and vice versa.}
    \label{tab:baselines}
    \vspace{-1em}
\end{table}

\subsection{Experimental Setup}
\noindent \textbf{Dataset.} For our experiments, we use the \emph{View-of-Delft} (VoD) dataset~\cite{palffy2022multi}, which provides synchronized and calibrated data captured by co-located sensors, including a 64-beam LiDAR, an RGB camera, RTK-GPS/IMU based odometer and a 4D radar sensor. As is often the case with datasets focused on object recognition, the test set annotations of the official VoD dataset are withheld for benchmarking. However, our task requires custom scene flow metrics for which we need annotations to generate ground truth labels. Therefore, we divide new splits ourselves from the official sets (\ie, \emph{Test}, \emph{Val}, \emph{Train}) to support our evaluation. 
Given sequences of data frames, we form pairs of consecutive radar point clouds as our scene flow samples for training and inference. We only generate ground truth scene flow and motion segmentation labels for samples from our  \emph{Val}, \emph{Test} sets and leave the \emph{Train} set unlabelled as our method has no need for ground truth scene flow annotations for training. 
Please see the supplementary details on our dataset separation and labelling process. 

\noindent \textbf{Metrics.} 
We use three standard metrics~\cite{liu2019flownet3d, mittal2020just, wu2020pointpwc} to evaluate different methods on scene flow estimation, including a) \emph{EPE} [m]: average end-point-error ($L_2$ distance) between ground truth scene flow vectors and predictions,
b) \emph{AccS/AccR}: the ratio of points that meet a strict/relaxed condition, \ie \emph{EPE} $<$ 0.05/0.1~m or the relative error $<$ 5\%/10\%. We also use the c) \emph{RNE} [m] metric~\cite{ding2022raflow} that computes resolution-normalized \emph{EPE} by dividing \emph{EPE} by the ratio of 4D radar and LiDAR resolution. This can induce sensor-specific consideration and maintain a fair comparison between different sensors. Besides, we compute the \emph{RNE} [m] for moving points and static points respectively, and denote them as \emph{MRNE} [m] and \emph{SRNE} [m]. 

\noindent \textbf{Baselines.}
For overall comparison, we apply
seven state-of-the-art methods as our baselines,
including five self-supervised learning-based methods~\cite{mittal2020just,wu2020pointpwc,kittenplon2021flowstep3d,baur2021slim,ding2022raflow} and two non-learning-based methods~\cite{besl1992icp,pontes2020scene}, as seen in~\cref{tab:baselines}. To keep a fair comparison, we use their default hyperparameter settings. We retrain scene flow models offline on our VoD \emph{Train} set for learning-based methods and directly optimize scene flow results online for non-learning-based ones.   

\noindent \textbf{Implementation details.}
We use the Adam optimizer~\cite{kingma2015adam} to train all models in our experiments. The learning rate is initially set as $0.001$ and exponentially decays by $0.9$ per epoch. The \emph{Val} set is used for both model selection during training and for determining the values of our hyperparameters. We set our classification threshold $\eta_b=0.5$ (c.f.~\cref{sec:model}) for all experiments. When activating our temporal update module, the length of mini-clips $T$ is set as 5. 
Please refer to the supplementary materials for our hyperparameter searching process.

\begin{table}[tbp!]
    \renewcommand\arraystretch{1}
    \setlength\tabcolsep{6pt}
    \scriptsize
    \centering
    \resizebox{0.9\columnwidth}{!}{%
    \begin{tabular}{@{}l|cccccccc@{}}
    \toprule
        & O & L & C & EPE [m]$\downarrow$  & AccS$\uparrow$  &  AccR$\uparrow$ & RNE [m]$\downarrow$  \\
    \midrule 
    (a) & & & & 0.228 & 0.184 & 0.392 & 0.091\\
    (b) & \checkmark &  &  & 0.161 & 0.203 & 0.442 & 0.065\\
    (c) & \checkmark & \checkmark &  & 0.145 & 0.228 & 0.482 & 0.058 \\
    (d) & \checkmark &  & \checkmark & 0.159 & 0.216 & 0.458 & 0.064 \\
    (e) & \checkmark & \checkmark & \checkmark & \textbf{0.141} & \textbf{0.233} & \textbf{0.499} & \textbf{0.057}\\
    \bottomrule
    \end{tabular}
    }
    \caption{Ablation experiments on combing supervision signals from diverse modalities. Abbreviations: odometer (O), LiDAR (L), camera (C). Note that we disable the temporal update scheme in this study to highlight the impact of modalities for training.}
    \label{tab:breakdown}
    \vspace{-1em}
\end{table}

\begin{figure}[tbp!]
    \centering
    \includegraphics[width=0.4\textwidth]{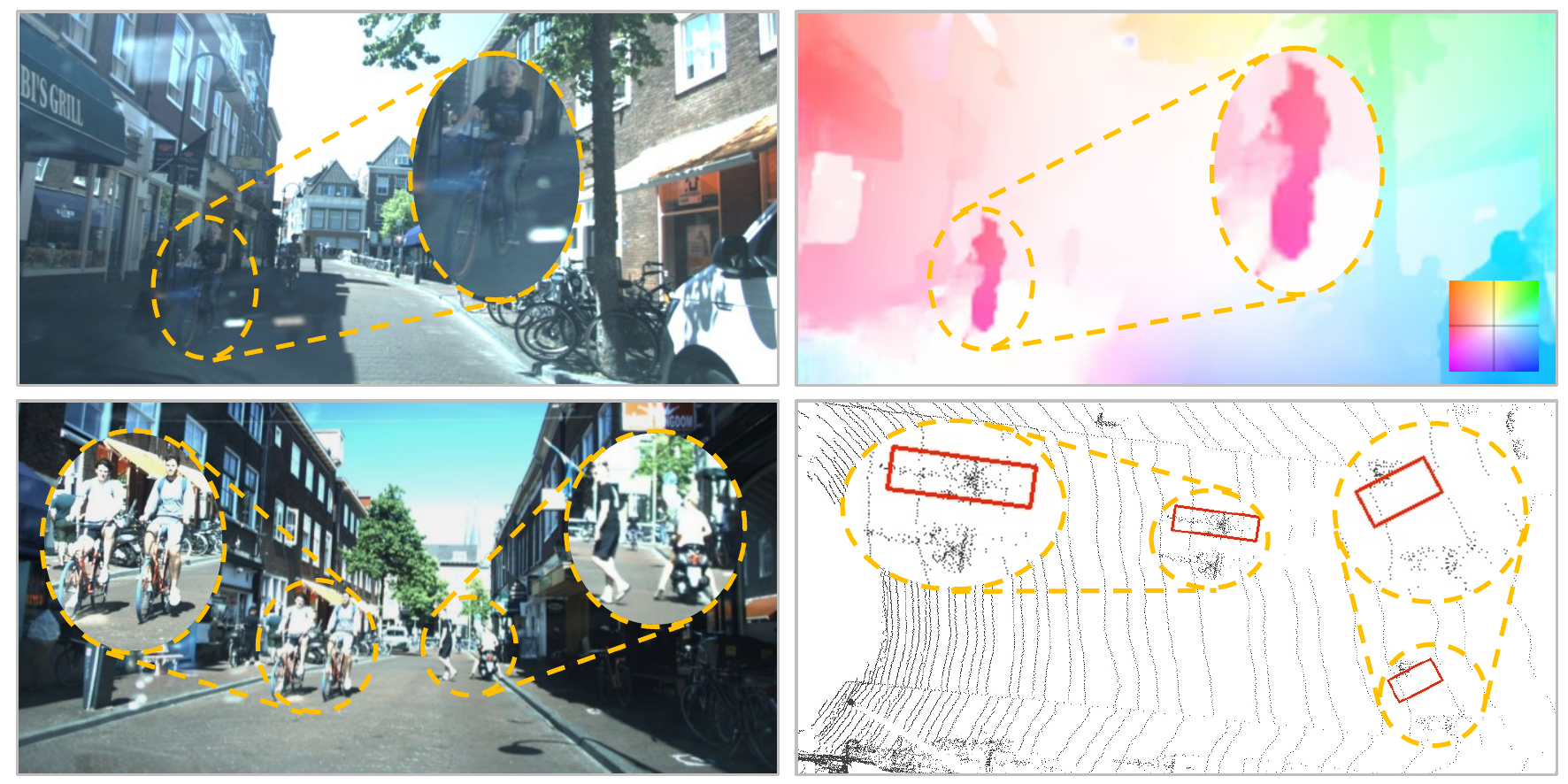}
    \caption{Illustration of the causes of noisy supervision signals from camera and LiDAR. The top row shows an example of the noisy optical flow estimation on RGB images. The bottom row exhibits unreliable object recognition on LiDAR point clouds. We enlarge regions of interest and mark them with \textcolor{myyellow}{amber} circles.}
    \label{fig:marginal}
    \vspace{-1.5em}
\end{figure}

\begin{figure*}[htbp!]
    \centering
    \includegraphics[width=0.925\textwidth]{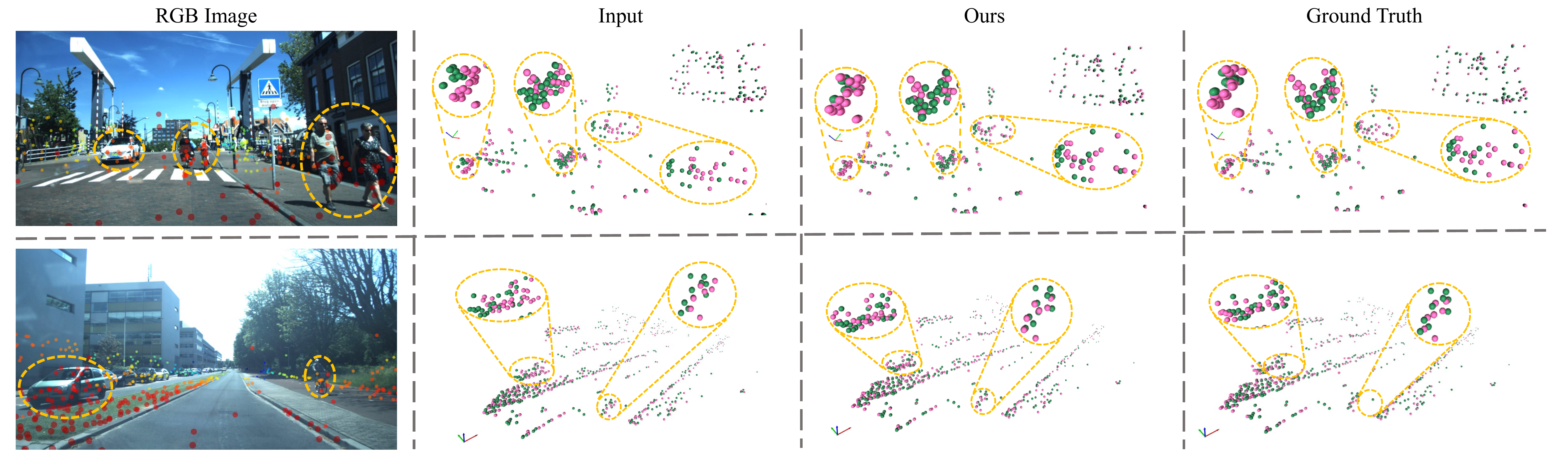}
    \caption{Qualitative scene flow results in two scenes. From left to right: 1) radar points from the source frame projected to the corresponding RGB image (points are coloured by distance from the sensor), 2) two input radar point clouds, the source one (\textcolor{mypink}{pink}) and the target one (\textcolor{mygreen}{green}), 3) the source point cloud warped by our predicted scene flow and the target radar point cloud, 4) the source point cloud warped by ground truth scene flow and the target one. We mark dynamic objects in \textcolor{myyellow}{amber} and apply the zooming-in operation for them. }
    \label{fig:vis}
    \vspace{-1em}
\end{figure*}

\subsection{Scene Flow Evaluation}

\noindent \textbf{Overall results.} We quantitatively compare the performance of our methods to baselines on the \emph{Test} set, as shown in~\cref{tab:baselines}. Compared with both non-learning-based and self-supervised counterparts, CMFlow shows remarkable improvement on all metrics by leveraging cross-modal supervision signals for training. Our method outperforms the second-best approach~\cite{ding2022raflow} by 37.6\% on EPE, implying drastically more reliability. 
Our performance is further improved when adding the temporal update scheme (c.f.~\cref{sec:model}) in the backbone, \ie CMFlow (T).
We also observe that the performance slightly degrades on the AccS metric when activating the temporal update. This suggests us that introducing temporal information will somewhat disturb the original features and thus reduce the fraction of already fine-grained (\eg EPE$ < $0.05 m) predictions. 

\noindent\textbf{Impact of modalities.} 
A key mission of this work is to investigate how supervision signals from different modalities help our radar scene flow estimation. To analyze their impact, we conduct ablation studies and show the breakdown results in~\cref{tab:breakdown}. 
The results show that the odometer leads to the biggest performance gain for our method (c.f. row (b)). This can be credited to the fact, that most measured points are static (\eg 90.5\% for the \emph{Test} set) and their flow predictions can be supervised well by the ego-motion supervision (\cref{eq:ego}). Moreover, the odometer guides the generation of pseudo motion segmentation labels (\cref{sec:loss}), which is indispensable for our two-stage model architecture. 

Based on row (b), both LiDAR and camera contribute to a further improvement on all metrics, as seen in {row (c)-(e)}. These results validate that our method can effectively exploit useful supervision signals from each modality to facilitate better scene flow estimation. However, compared to that from the odometer, the gains brought by these two sensors are smaller, especially for the camera, which only increases the AccS by 1.3\%. In our opinion, the reason for this is two-fold. First, in~\cref{loss:mot} and~\cref{loss:opt}, the pseudo scene/optical flow labels only constrain the flow vectors of identified moving points, which are significantly fewer than static ones. Thus, they have limited influence on the overall performance. Second, the supervision signals extracted from these sensor modalities could be noisy, as exhibited in~\cref{fig:marginal}. The dashboard reflections on the car windscreen severely disturb the optical flow estimation on images, which further results in erroneous supervision in~\cref{loss:opt}. As for LiDAR point clouds, both inaccurate object localization and false negative detections occur in the 3D MOT results, which not only affect the generation of pseudo motion segmentation labels (\cref{eq:seg}) but bring incorrect constraints to scene flow in~\cref{loss:mot}. 

\begin{figure}[tbp!]
    \centering
    \includegraphics[width=0.475\textwidth]{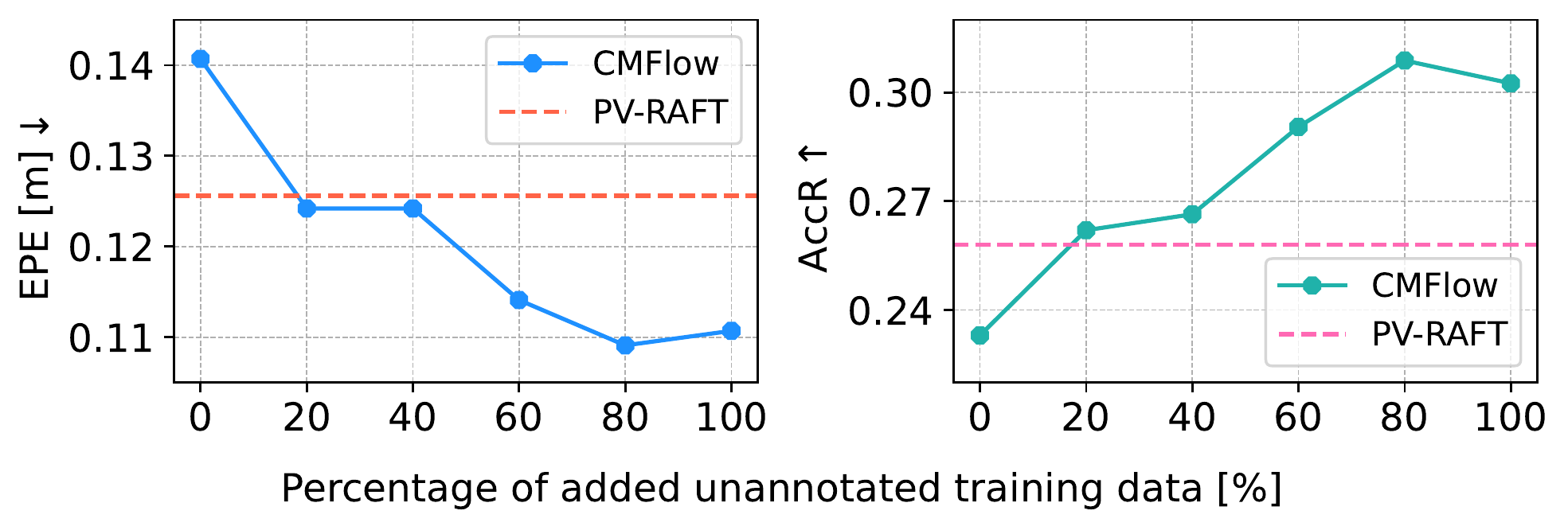}
    \caption{Analysis of the performance when adding more unannotated training data. For the training of CMFlow, we retain the samples from the \emph{Train} set and add different percentages of data from the extra unannotated VoD part, which provides $\sim28.5$k more training samples.} 
    \label{fig:unanno}
    \vspace{-2em}
\end{figure}

\noindent\textbf{Impact of the amount of unannotated data.}
Being able to use more unannotated data for training is an inherent advantage of our cross-modal supervised method compared to fully-supervised ones as no additional annotation efforts are required. Here, we are interested in if the performance of CMFlow could surpass that of fully-supervised methods when more unannotated data is available for training. To this end, we further mix in an extra amount of unlabeled data provided by the VoD dataset~\cite{palffy2022multi}\footnote{This part is currently in a beta testing phase, only available for selected research groups. Except for having no object annotations, it provides the same modalities of input data as the official one and has $\sim28.5$k frames.} in our cross-modal training sets. As for fully-supervised methods, we select the state-of-the-art method, PV-RAFT~\cite{wei2021pv}, for comparison. As PV-RAFT needs ground truth scene flow annotations for training, we utilize all available annotated samples from the \emph{Train} set for it. The analysis of the correlation between the percentage of added unannotated training data and the performance is shown in~\cref{fig:unanno}.  As we can see, the performance of CMFlow improves by a large margin on both two metrics by using extra training data. In particular, after adding only 20$\%$ of the extra unannotated training samples ($\sim$140\% more than the number of training samples used for PV-RAFT), CMFlow can already outperform PV-RAFT~\cite{wei2021pv} trained with less annotated samples. This implies the promise of our method for utilizing a large amount of unannotated data in the wild.
\begin{table}[tbp!]
    \renewcommand\arraystretch{1}
    \setlength\tabcolsep{6pt}
    \footnotesize
    \centering
    \resizebox{0.85\columnwidth}{!}{%
    \begin{tabular}{@{}l|ccccc@{}}
    \toprule
        & Label $\mathbf{S}^{v}$ &  Label $\mathbf{S}^{l}$ & A.D.  & mIoU (\%) & Gain (\%)\\
    \midrule 
    (a) & & & &  46.9 & - \\
    (b) & \checkmark & &  & 52.8 & {+5.9} \\
    (c) & \checkmark &  \checkmark &  & 54.1 & +1.3\\
    (d) & \checkmark &  \checkmark & \checkmark &  57.1 & +3.0\\
    \bottomrule
    \end{tabular}
    }
    \caption{Motion segmentation evaluation. For row (a), we report the results of the self-supervised baseline~\cite{ding2022raflow}. In row (b), we replace $\mathbf{S}$ by $\mathbf{S}^{v}$ provided by odometer. A.D. denotes adding extra unannotated data (c.f.~\cref{fig:unanno}) for training.}
    \label{tab:seg}
    \vspace{-1.5em}
\end{table}

\noindent\textbf{Qualitative results.}
In~\cref{fig:vis}, we showcase example scene flow results of CMFlow (trained with the extra training samples) compared to the ground truth scene flow. By applying the estimated scene flow (\ie moving each point in the source frame by its estimated 3D motion vector), both the static background and multiple objects with different motion patterns are aligned well between two frames. It can also be observed that our method can give accurate scene flow predictions, close to the ground truth one.

\begin{figure}[tbp!]
    \centering
    \includegraphics[width=0.485\textwidth]{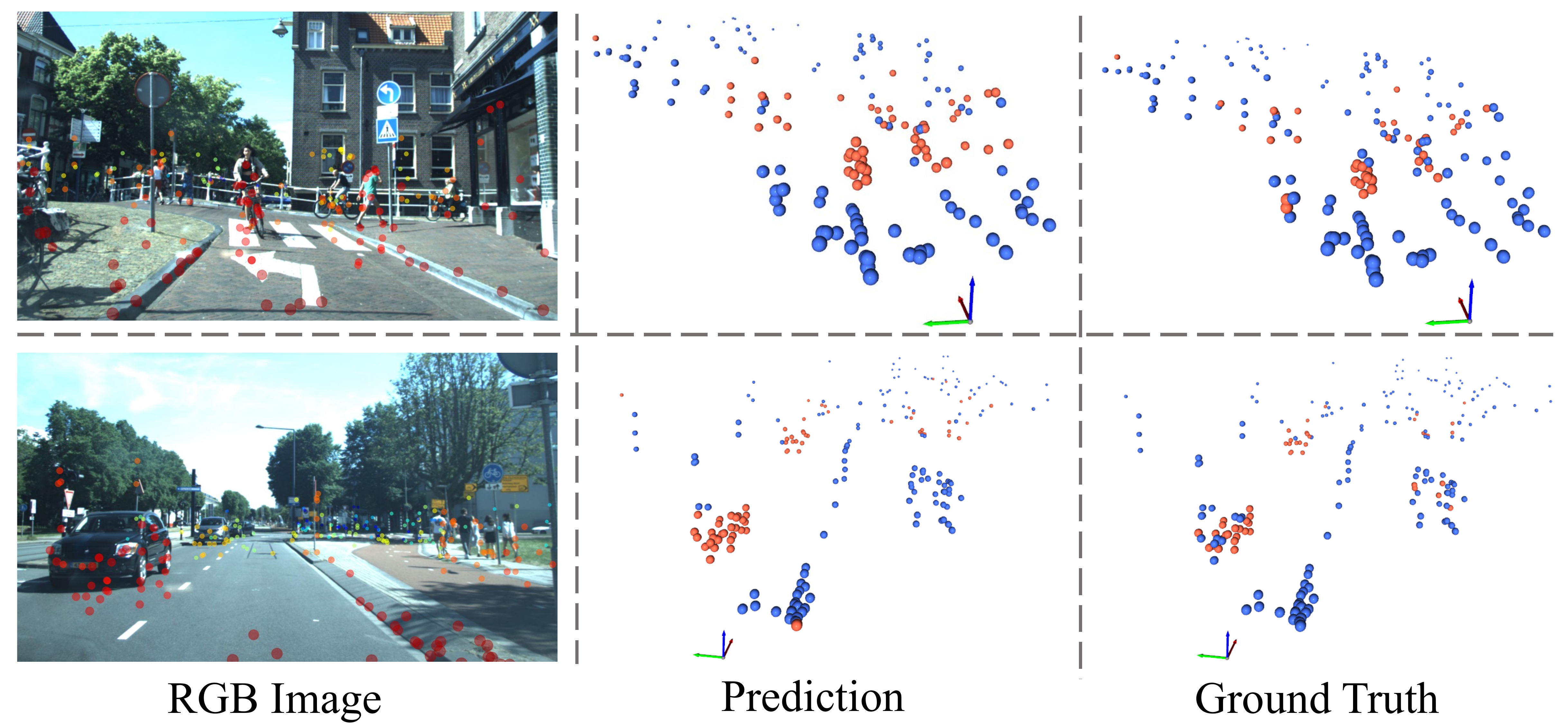}
    \caption{Visualization of motion segmentation. The left column shows the corresponding image with radar points (coloured by range) projected onto it. In the middle and right columns, moving points are shown in \textcolor{myorange}{orange} while static points are shown in \textcolor{myblue}{blue}.\looseness = -1}
    \label{fig:seg}
    \vspace{-1.5em}
\end{figure}

\subsection{Subtask Evaluation}
\noindent\textbf{Motion segmentation evaluation.} Apart from scene flow estimation, our multi-task model can additionally predict a motion segmentation mask that represents the real moving status of each radar point (c.f.~\cref{sec:model}). Here, we evaluate this prediction of CMFlow and analyze the impact of its performance in~\cref{tab:seg}. 
Since this is a binary classification task for each point, the mean intersection over union (mIoU) is computed by taking the IoU of moving and static classes and averaging them. As the two ingredients to form the final pseudo motion segmentation label $\mathbf{S}$ used to supervise our output in~\cref{eq:seg}, both $\mathbf{S}^{v}$ and $\mathbf{S}^{l}$ contributes to our performance improvement on the motion segmentation task (row (a)-(c)). Moreover, our mIoU further increases with extra training data to learn motion segmentation on 4D radar point clouds. We also visualize some qualitative motion segmentation results of row (d) in~\cref{fig:seg}, where our method can segment moving points belonging to multiple dynamic objects accurately in complicated scenarios. 

\begin{figure}[tbp!]
    \centering
    \includegraphics[width=0.45\textwidth]{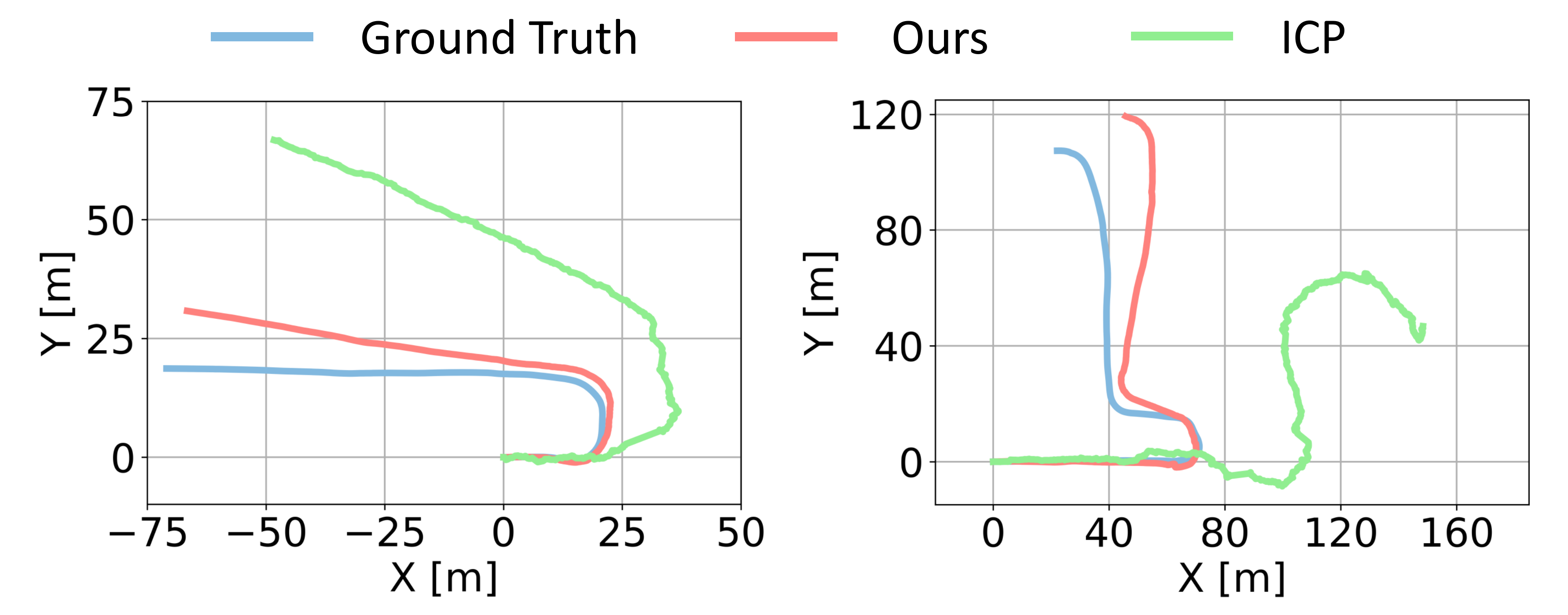}
    \caption{Odometry results as a byproduct of our scene flow estimation. The ground truth is generated using the RTK-GPS/IMU measurements.  We plot the results on two challenging test sequences. Please refer to more results in supplementary videos.} 
    \label{fig:ego}
    \vspace{-1.5em}
\end{figure}

\noindent\textbf{Ego-motion estimation evaluation.}
One important feature of our method is that we can estimate a rigid transformation that represents relative radar ego-motion transform between two consecutive frames in dynamic scenes (c.f.~\cref{sec:model}). To demonstrate this feature, we evaluate this output on the \emph{Test} set and show the ablation study results in~\cref{tab:ego} with two metrics: the relative translation error (RTE) and the relative angular error (RAE). With the cross-modal supervision from the odometry in row (b), we can directly constrain our ego-motion estimation in~\cref{eq:ego} and thus improve the performance by a large margin. Using LiDAR and camera supervision (c.f. row (c)) can also help as they lead to better motion segmentation and scene flow outputs, which further benefit the compactly associated ego-motion estimation. We also activate the temporal update module in the backbone, which also increases the overall performance. 
With our high-level accuracy on ego-motion estimation between consecutive frames, we are also interested in whether our results can be used for the more challenging long-term odometry task. We accumulate the inter-frame transformation estimations and plot two ego-vehicle trajectories in~\cref{fig:ego}. Without any global optimization, our method can provide accurate long-term trajectory estimation in dynamic scenes by only estimating inter-frame transformations and remarkably outperform the ICP baseline~\cite{besl1992icp}. 

\begin{table}[tbp!]
    \renewcommand\arraystretch{1}
    \setlength\tabcolsep{6.5pt}
    \footnotesize
    \centering
    \resizebox{0.8\columnwidth}{!}{%
    \begin{tabular}{@{}l|ccccccc@{}}
    \toprule
      & O & L + C & A.D. & T &  {RTE [m]} & {RAE [$^\circ$]}\\
     \midrule
      (a) & &  & & & 0.090 & 0.336 \\
      (b) & \checkmark & &  & &  0.086 & 0.183\\
      (c) & \checkmark &  \checkmark & & &  0.085 & 0.145 \\
      (d) & \checkmark & \checkmark & \checkmark & & 
      0.071 &	\textbf{0.089} \\
      (e) & \checkmark & \checkmark & \checkmark & \checkmark & 
      \textbf{0.066} & 0.090 \\
    \bottomrule
    \end{tabular}
    }
    \caption{Evaluation of ego-motion estimation between two frames.  T denotes we activate the temporal update module.} 
    \label{tab:ego}
    \vspace{-1.5em}
\end{table}

\vspace{-1em}
\section{Conclusion}
\vspace{-0.5em}
In this paper, we presented a novel cross-modal supervised approach, CMFlow, for estimating 4D radar scene flows. CMFlow is unique in that it does not require manual annotation for training. Instead, it uses complementary supervision signals  retrieved from co-located heterogeneous sensors, such as odometer, LiDAR and camera, to constrain the outputs from our multi-task model.
Our experiments show that CMFlow outperforms our baseline methods in all metrics and can surpass the fully-supervised method when sufficient unannotated samples are used in our training. 
CMFlow can also improve two downstream tasks, \ie, motion segmentation and ego-motion estimation. 
We hope our work will inspire further investigation of cross-modal supervision for scene flow estimation and its application to more downstream tasks, such as multi-object tracking and point cloud accumulation.

\vspace{0.5em}
\small{
\noindent\textbf{Acknowledgment.}}
\footnotesize{This research is supported by the EPSRC, as part
of the CDT in Robotics and Autonomous Systems hosted at the Edinburgh
Centre of Robotics (EP/S023208/1), and the Google Cloud Research Cred-
its program with the award GCP19980904.}


{\small
\bibliographystyle{ieee_fullname}
\bibliography{egbib}
}

\end{document}